\documentclass[final]{iasart}			
\usepackage[pdftex]{graphicx}
\graphicspath{{./Fig/}}
\DeclareGraphicsExtensions{.pdf,.jpg,.png}
\usepackage[utf8]{inputenc}
\usepackage[T1]{fontenc}

\usepackage{color} 

\usepackage{natbib}

\usepackage{amsfonts}
\usepackage{amsmath,bm}
\usepackage{amsthm} 

\usepackage{mathrsfs}
\usepackage{eucal}
\usepackage{scalerel}

\usepackage{array}

\usepackage{url}
\usepackage[hidelinks]{hyperref}

\usepackage[caption=false,font=footnotesize]{subfig}

\usepackage{stfloats}

\usepackage{paralist}

\usepackage[all]{nowidow}

\newcommand\scale[2]{\vstretch{#1}{\hstretch{#1}{#2}}}
	
\newcommand{\LP}{\if@draft
		\mathbin{\ooalign{$\bigtriangleup$\crcr\hidewidth \raise.14em\hbox{$\scale{0.7}{\scriptscriptstyle+}$}\hidewidth}}
	\else
		\mathbin{\mathpalette\LIPcls+}
	\fi}
\newcommand{\LM}{\if@draft
		\mathbin{\ooalign{$\bigtriangleup$\crcr\hidewidth \raise.14em\hbox{$\scale{0.7}{\scriptscriptstyle-}$}\hidewidth}}
	\else
		\mathbin{\mathpalette\LIPcls-}
	\fi}
\newcommand{\LT}{\if@draft
	  \mathbin{\ooalign{$\bigtriangleup$\crcr\hidewidth \raise.14em\hbox{$\scale{0.7}{\scriptscriptstyle\times}$}\hidewidth}}
	\else
		\mathbin{\mathpalette\LIPcls\times}
	\fi}

\newcommand{\LIPcls}[2]{%
  \ooalign{$#1\bigtriangleup$\crcr  \hidewidth\raisefix{#1}\hbox{$#1\scale{0.45}{\bm{#2}}$}\hidewidth}}

\def\raisefix#1{%
  \ifx#1\displaystyle
    \raise.14em
  \else
    \ifx#1\textstyle
      \raise.14em
    \else
      \ifx#1\scriptstyle
        \raise.112em
      \else
        \raise.0933em
      \fi
    \fi
  \fi
}

\newcommand{\Real}{\mathbb R}

\newcommand{\la}{\lambda}

\theoremstyle{remark}
\newtheorem*{remark}{Remark}

\def\citeapos#1{\citeauthor{#1}'s (\citeyear{#1})}

\newif\ifcompile
\compiletrue 

\newcommand{\myinclude}[1]{\if@draft
	\include{#1}
\else
  \input{#1}
\fi}

\title{Region homogeneity in the Logarithmic Image Processing framework: application to region growing algorithms}
\shorttitle{LIP Region Homogeneity}
\shortauthors{Noyel G \etal}

\author[1,2]{Guillaume Noyel}
\author[3,1]{Michel Jourlin}

\email{guillaume.noyel@i-pri.org,
michel.jourlin@univ-st-etienne.fr}

\affiliation[1]{International Prevention Research Institute, Lyon , France}

\affiliation[2]{University of Strathclyde Institute of Global Public Health, Dardilly - Lyon Ouest, France}

\affiliation[3]{Laboratoire Hubert Curien, UMR CNRS 5516, Universit\'e Jean Monnet, Saint-Etienne, France}

\abstract{In order to create an image segmentation method robust to lighting changes, two novel homogeneity criteria of an image region were studied. Both were defined using the Logarithmic Image Processing (LIP) framework whose laws model lighting changes. The first criterion estimates the LIP-additive homogeneity and is based on the LIP-additive law. It is theoretically insensitive to lighting changes caused by variations of the camera exposure-time or source intensity. The second, the LIP-multiplicative homogeneity criterion, is based on the LIP-multiplicative law and is insensitive to changes due to variations of the object thickness or opacity. Each criterion is then applied in \citeapos{Revol1997} region growing method which is based on the homogeneity of an image region. The region growing method becomes therefore robust to the lighting changes specific to each criterion. Experiments on simulated and on real images presenting lighting variations prove the robustness of the criteria to those variations. Compared to a state-of the art method based on the image component-tree, ours is more robust. These results open the way to numerous applications where the lighting is uncontrolled or partially controlled.}


\begin{document}
\begin{paper}

%
%
\section{Introduction}


Segmentation of images acquired with different lighting conditions is a challenging task in image analysis that can occur in many settings such as visual inspection for industry \citep{Cord2010,Noyel2011,Noyel2013,Parradenis2011}, medical images \citep{Noyel2014,Noyel2017c}, security \citep{Foresti2005}, driving assistance \citep{Hautiere2006}, etc. 
Lighting variations obviously represent a substantial obstacle for the development of reliable image processing algorithms. To overcome this difficulty, we can determine the two following approaches: 

\begin{inparaenum}[(i)]
\item Firstly, a \textit{practical} approach consisting of circumventing the problem, e.g. by a learning process. This is often the case for face recognition, where a database is created by varying together the acquisition angle and the intensity of the lighting source \citep{Ramaiah2015}. An alternative way to learning processes aims at extracting from images some characteristics independent of lighting conditions \citep{Wang2004,Shah2015}. Another alternative aims at attenuating the lighting variations effects by means of various methods like shadow suppression \citep{Zhang2018}.

\item Secondly, a \textit{fundamental} approach consists of creating models and algorithms almost insensitive to such variations. In that respect, \citet{Chen2006} define the Logarithmic Total Variation (LTV) model in order to remove varying illumination in face images. \citet{Yu2017} regard an image in the three-dimensional space as a grey wave composed of peaks and troughs which divide the image into many local sub-regions. The \textit{Retinex} model takes into account the human perception of colours and contrasts. It has been extensively reviewed by \citet{Elad2003} alongside some related illumination compensation methods. We can also mention the representation methods of the image level sets by trees of their connected components, such as: the \textit{component-tree}, also named \textit{max-tree} \citep{Salembier1998}, or the \textit{tree of shapes}, also named \textit{inclusion tree} \citep{Monasse2000}. They are invariant to contrast changes due to a continuous and increasing function applied to the image \citep{Monasse2000}. Recently, \citet{Passat2011} introduced a segmentation method by \textit{component-tree} presenting a certain robustness to intensity inhomogeneity provided that the maximal image contrast is in an intermediate range: not too low and not too huge. The connected filtering on tree-based shape spaces of \cite{Xu2016} is also theoretically invariant to those contrast changes.
\end{inparaenum}

Nevertheless, only a few papers focus on the causes of lighting variations. For this purpose, \citet{Jourlin2018} have defined two new homogeneity criteria in the LIP (Logarithmic Image Processing) framework, namely the LIP-additive and LIP-multiplicative homogeneity criteria. Each criterion is based either on the LIP law of addition $\LP$ of two images or the multiplication $\LT$ of an image by a scalar. Such laws are now well-known to model the lighting variations affecting the image acquisition step \citep{Jourlin2016}. The LIP-additive law $\LP$ especially models changes of source intensity or exposure-time whereas the LIP-multiplicative law $\LT$ models the thickness (or opacity) changes of the observed object.

The aim of this paper is to study these homogeneity criteria in details. At the theoretical level, we  will demonstrate their insensitivity to the lighting variations which are modelled by the corresponding LIP-laws.  At the application level, we will prove their robustness to lighting changes on both simulated and on real images. For segmentation purpose we will apply them within the \citeapos{Revol1997} algorithm which is in the class of \textit{Region Growing} methods. This algorithm has the specificity of evaluating the homogeneity of an image region at each step. Its segmentation results will be compared to those obtained by the component-tree method of \citet{Passat2011}.

%
%
\section{Materials and methods}
\label{sec:mat_meth}

In this section, we will present the theoretical methods and their validation by the experimental material. First, we will summarise the Logarithmic Image Processing model. Second, we will present the specific lighting variations which are modelled by each LIP law. Third, we will introduce the LIP-additive and LIP-multiplicative homogeneity criteria. Fourth, we will recall the \citeapos{Revol1997} region growing algorithm based on the homogeneity of an image region. Fifth, for comparison purpose, we will recall the sate-of-the-art method of \citet{Passat2011}. Finally, we will present the experimental validation with simulated and real images.

%
%

\subsection{Logarithmic Image Processing}
\label{ssec:mat_meth:LIP}

Introduced by \citet{Jourlin1988,Jourlin2001}, the LIP model possesses strong physical properties. Let $f$ and $g$ be two grey level functions defined on the spatial support $D \subset \Real^n$, with values in the grey scale $[0,M[$ where $M \in \Real$. In fact, the addition $f \LP g$  of $f$ and $g$ and the scalar multiplication $\la \LT f$ of $f$ by a real number $\la$ are both deduced from the optical \textit{Transmittance Law} according to:
\begin{align}
	f \LP g &= f + g - fg/M, \label{eq:LIP:plus}\\%
	\lambda \LT f &= M - M \left( 1 - f/M \right)^{\lambda}. \label{eq:LIP:times}%
\end{align}

Let us note that in the LIP framework, the grey scale $[0,M[$ is inverted in comparison with the classical one. Indeed, due to the transmittance law, $0$ represents the ``white'' extremity of the scale, which corresponds to the observation of the source intensity. The reason of such an inversion is that $0$ appears as the neutral element of the addition $\LP$, i.e. when no obstacle is placed between the source and the sensor.

From equation \ref{eq:LIP:plus}, the subtraction between two grey level functions is easily deduced:

\begin{align}
	f \LM g &= \frac{f-g}{1-\frac{g}{M}}. \label{eq:LIP:minus}%
\end{align}

\begin{remark}
$f \LM g$ can take negative values except for $f(x)\geq g(x)$ at each point $x$. In such a case, the subtraction $f \LM g$ is a grey level image and the difference $f(x) \LM g(x)$ represents the \textit{Logarithmic Additive Contrast} between the grey levels $f(x)$ and $g(x)$ \citep{Jourlin2001}. This difference is the grey level which must be added to the brightest one (here $g(x)$) in order to obtain the darkest one (here $f(x)$).
\end{remark}

\begin{remark}
Let us note that for 8-bit digitised images, $M$ is equal to 256 and the grey levels vary from 0 to 255.
\end{remark}

At the mathematical level, it has been demonstrated \citep{Jourlin2001} that the laws $\LP$ and $\LT$ give a \textit{Vector Space} structure to the set $I(D,[0,M[)$ of images defined on a same spatial support $D$. This allows the use of many mathematical tools specific to this kind of space. In order to complete the outstanding properties of the LIP Model, let us recall that \citet{Brailean1991} have established its consistency with the Human Visual System, which opens the way to process images as a human eye would do.

%
%

\subsection{Which lighting variations are modelled by each LIP law?}
\label{ssec:mat_meth:light_var}

\citet{Carre2014} and \citet{Deshayes2015} have shown that the LIP-addition (respectively subtraction) of a constant to an image (resp. from an image) perfectly models a decrease (resp. an increase) of the camera exposure-time or of the source intensity.
By definition of the scalar multiplicative law, the scalar $\la$ appears as a ``thickness'' parameter: in fact, if $f$ is a grey level image, one can associate to it a virtual half-transparent object producing $f$ in transmission. $\la \LT f$ represents then the image we would obtain by stacking $f$ on itself $\la$ times. The scalar multiplicative law models therefore the ``opacity'' changing of a half-transparent object or the thickness changing of a homogeneous object.

%
%

\subsection{Homogeneity criteria: LIP-additive homogeneity and LIP-multiplicative homogeneity}
\label{ssec:mat_meth:hom_crt}

Let $f$ be a grey level function defined on the spatial support $D \subset \Real^n$, with values in the grey scale $[0,M[$ and consider a region $R$ of $D$. There exist some classical parameters evaluating the homogeneity of the image $f$ in $R$, like the standard deviation $\sigma_f(R)$ or the ``diameter'' of $R$ in the mathematical sense defined by $\sup_{x \in R}{f(x)} - \inf_{x \in R}{f(x)}$ which is nothing but the dynamic of the image region. Such homogeneity parameters are obviously sensitive to lighting variations, which justifies the introduction of the two new criteria: the LIP-additive and the LIP-multiplicative homogeneity.

\begin{remark}
When there is no ambiguity, we will consider that the homogeneity of an image $f$ over a region $R$ is equivalent to the homogeneity of the region $R$. The region $R$ is indeed supposed to be associated to the image $f$.
\end{remark}

%

\subsubsection{LIP-additive homogeneity}
\label{ssec:mat_meth:hom_crt:LIPadd}

The homogeneity $H^{\LP}_{f}(R)$ of the region $R$ is defined as its dynamic in the LIP sense:
\begin{align}
H^{\LP}_{f}(R) &= \sup_{x \in R}{f(x)} \LM \inf_{x \in R}{f(x)} \label{eq:LIPadd_H}
\end{align}

Let us note that $H^{\LP}_{f}(R)$ lies in $[0,M[$ and represents thus a grey level which can be interpreted as the maximal \textit{Logarithmic Additive Contrast} observed in the region \citep{Jourlin2001}.

The most important property of this homogeneity criterion is its insensitivity to exposure time variations modelled by LIP-addition/subtraction of a constant $C$:

\begin{align}
H^{\LP}_{f \LP C}(R) &= H^{\LP}_{f}(R) \label{eq:LIPadd_H:ist}
\end{align}

\begin{proof}
Let us first remark that 
\begin{align*}
\sup_{x \in R}{\{(f \LP C)(x)\}} &= \sup_{x \in R}{\{f(x)\}} \LP C.
\end{align*}
In fact, equation \ref{eq:LIP:plus} gives: 
\begin{align*}
& \sup{\{f\}} \LP C  = \sup{\{f\}} + C - (C.\sup{\{f\}})/M \nonumber\\
&= \left(1-C/M\right) \sup{\{f\}} + C =\sup{\left\{ f \left(1-C/M \right) \right\}} + C \nonumber\\
&= \sup{\left\{ f \left(1-C/M \right) + C \right\}} = \sup{\left\{ f + C - f.C/M \right\}} \nonumber\\
&= \sup{\left\{ f \LP C \right\}}.
\end{align*}

In the same way, we have:
\\$\inf_{x \in R}{\{(f \LP C)(x)\}} = \inf_{x \in R}{\{f(x)\}} \LP C$.

In such conditions, equation \ref{eq:LIPadd_H} can be written:
\begin{align*}
H^{\LP}_{f \LP C}(R) &= (\sup_{x \in R}{\{f(x)\}} \LP C) \LM (\inf_{x \in R}{\{f(x)\}} \LP C).
\end{align*}
The following computation shows that $(A \LP C) \LM (B \LP C)$ is equal to $A \LM B$:

\begin{align*}
&(A \LP C) \LM (B \LP C) = \frac{ A+C-\frac{A.C}{M} - \left( B+C- \frac{B.C}{M} \right) }{ 1 - \frac{ B+C- \frac{B.C}{M} }{M} }\\
&= \frac{ (A-B)\left(1-\frac{C}{M} \right) }{ 1-\frac{B}{M} - \frac{C}{M} - \frac{B.C}{M^2} }  
=  \frac{ (A-B)\left(1-\frac{C}{M} \right) }{ \left(1-\frac{B}{M}\right) \left( 1 - \frac{C}{M} \right) }
=  \frac{ (A-B) }{ \left(1-\frac{B}{M}\right) }\\
&= A \LM B.
\end{align*}
The two previous results prove equation \ref{eq:LIPadd_H:ist}.
\end{proof}


\subsubsection{LIP-multiplicative homogeneity}
\label{ssec:mat_meth:hom_crt:LIPmult}

Now, we propose a homogeneity criterion based on the LIP scalar multiplication and more precisely on the notion of \textit{Logarithmic Multiplicative Contrast} (LMC) \citep{Jourlin2012}. Let us recall that the LMC associated to a pair of grey levels $g_1$ and $g_2$, where $g_1$ and $g_2$ are strictly positive, is the real number defined by the formula:

\begin{align}
LMC(g_1,g_2) \LT \min{(g_1,g_2)} &= \max{(g_1,g_2)}. \label{eq:LMC:intro} 
\end{align}

In the context of images acquired in transmission, this means that the LMC of two grey levels represents the number of times the brightest of them must be stacked upon itself to get the darkest one.
In such conditions, the LIP multiplicative homogeneity of a region $R$ is computed according to:

\begin{align}
H^{\LT}_{f}(R) &= LMC (\sup_{x \in R}{\{f(x)\}} , \inf_{x \in R}{\{f(x)\}} ) \label{eq:LIPmult_H} \\
&=\frac{\ln{\left(1- \sup_{x \in R}{\{f(x)\}}/ M \right)}}{\ln{\left(1- \inf_{x \in R}{\{f(x)\}}/M \right)}}. \label{eq:LIPmult_H2} 
\end{align}

\begin{remark} 
In the case where $\inf_{x \in R} f(x)=0$, one can replace this value by $1$ in order to avoid an infinite value for $H^{\LT}_{f}(R)$.
\end{remark} 
This homogeneity criterion possesses the fundamental property to be insensitive to variations of lighting modelled by the LIP-multiplicative law $\LT$, like opacity or thickness variations:

\begin{align}
H^{\LT}_{\la \LT f}(R) &= H^{\LT}_{f}(R). \label{eq:LIPmult_H:ist}
\end{align}

\begin{proof}
Suppose that $H_f^{\LT} (R)=\mu$, which means that $\sup_{x \in R}{f(x)} = \mu \LT \inf_{x \in R}{f(x)}$.
We have to prove the equality: 
\begin{align}
\sup_{x \in R} \la \LT f(x) &= \mu \LT \inf_{x \in R} \la \LT f(x). \label{eq:proof:ist_LIPmult_H:1}
\end{align}

Given $\la \geq 0$, it is easy to establish that:
\begin{align*}
\sup_{x \in R}{\{ \la \LT f(x) \}} &= \sup_{x \in R}{\{ M(1-(1-f(x)/M)^{\la}) \}} \nonumber\\
&= M(1-\inf_{x \in R}{\{(1-f(x)/M)^{\la}\}})  \nonumber\\
&= M(1-( \inf_{x \in R}{ \left\{ 1-f(x)/M \right\} })^{\la})  \nonumber\\
&= M(1-(1-\sup_{x \in R}{\{f(x)\}}/M)^{\la})  \nonumber\\
&= \la \LT \sup_{x \in R} f(x). \label{eq:proof:ist_LIPmult_H:x}
\end{align*}

In the same way,  $\inf_{x \in R} \la \LT f(x)= \la \LT \inf_{x \in R} f(x)$.
Thus, equation \ref{eq:proof:ist_LIPmult_H:1} is established, which ends the proof. 

\end{proof}

\subsection{A region growing algorithm based on region homogeneity}
\label{ssec:mat_meth:Revol}

\citet{Revol1997} presented a region growing method which minimises the region variance. In the sequel, we propose to replace the variance by one of the homogeneity criteria $H^{\LP}$ or $H^{\LT}$. Let us present the principles of this algorithm. The details are given in \citep{Revol1997}.

Let $dyn_f(R)$ be the dynamic range of the image $f$ over the region $R$. It is defined by $dyn_f(R) = [\min_{x \in R}\{f(x)\} , \max_{x \in R}\{f(x)\}]$.

At each step $n$, a region $R_n$ is dilated by a unitary structuring element $N$, e.g a square of length side 3 pixels \citep{Minkowski1903,Matheron1967,Serra1982,Najman2013}. This gives a region $D_{n+1}=R_n \oplus N$ whose homogeneity $H_f(D_{n+1})$ is measured. Depending on a threshold value $t$, the region is considered as homogeneous or not.
		\begin{enumerate}
			\item If $H_f(D_{n+1}) \leq t$, the region $D_{n+1}$ is then considered as homogeneous. The new region $R_{n+1}$ becomes $D_{n+1}$.
			
			\item If $H_f(D_{n+1}) > t$, the region $D_{n+1}$ is then inhomogeneous. A \textit{reduction} process $Rdct_f(D_{n+1})$ truncates the extremal classes of the histogram of $D_{n+1}$  by keeping only the pixels belonging to the dynamic range of $R_n$, $dyn_f(Rdct_f(D_{n+1}))=dyn_f(R_n)$. This gives a region $D'_{n+1}$.
				\begin{enumerate}[2.a ]
					\item If $H_f(D'_{n+1}) \leq t$, the region $D'_{n+1}$ is then modified by an \textit{extension} process which adds the neighbouring pixels to the region according to both  following conditions. i) The new pixel values have a LIP-difference less or equal than 1 with the maximum or the minimum values of $f$ over $D'_{n+1}$. ii) The new pixels give an homogeneous region. The new region $R_{n+1}$ becomes the extension of $D'_{n+1}$: $E(D'_{n+1})$.
					
					\item If $H_f(D'_{n+1}) > t$, a \textit{contraction} process reduces the dynamic range of the region $D'_{n+1}$ by removing the extremal classes of its histogram until the region becomes homogeneous. LIP-addition and LIP-subtraction replace the classical addition and subtraction used by \citet{Revol1997}. The new region $R_{n+1}$ is the contraction of $D'_{n+1}$: $Ctrct(D'_{n+1})$.
				\end{enumerate}
		\end{enumerate}
The algorithm stops when the region $R_{n+1}$ does not grow any more.

For a fully automatic segmentation method, the seed points could be chosen as the minima of a function such as: a gradient of the image - like for the watershed transform \citep{Beucher1992} - or a map of Asplund's distances of the image \citep{Jourlin2012,Jourlin2014,Noyel2015,Noyel2016,Noyel2017a}. The seed points could also be automatically determined by a previous classification \citep{Noyel2007,Noyel2010,Noyel2014}. As the homogeneity criterion is invariant to contrast changes with an optical cause, the threshold could be learnt from a training database \citep{Lu2017}. It could also be automatically chosen as the most frequent contrast on every pixel neighbourhoods \citep{Jourlin2016}.

%
%

\subsection{A state of the art method based on the image component-tree}
\label{ssec:mat_meth:Passat}

Our method will be compared to the state-of-the-art one of \citet{Passat2011} which is based on level sets and presents a certain robustness to lighting variations. It consists of an interactive segmentation by component-tree.  A component-tree is the tree formed by the inclusion relation between the binary components of the successive image level sets. A node $N$ of the component-tree is a connected component of the image level set $X_v(f)=\{ x \in D | f(x) \geq v \}$.
Let $\mathcal{K}$ be the set of all connected components of all level sets. 
The subset $\mathcal{K}' \subset \mathcal{K}$ of the connected components which best fits the user selected region $G$ in $D$ is automatically determined by the following minimisation problem: 
\begin{equation*}
\widehat{\mathcal{K}} = \arg \min_{\mathcal{K' \in \mathcal{P(\mathcal{K})}}} \{ d^{\alpha}( \cup_{N \in \mathcal{K}'} N , G ) \}.
\label{eq:mini:Passat}
\end{equation*} 
$\mathcal{P(\mathcal{K})}$ is the set of the parts of $\mathcal{K}$. The pseudo-distance $d^{\alpha}$ is defined by 
\begin{equation*}
d^{\alpha}(X,Y) = \alpha |X \setminus Y | + (1-\alpha) |Y \setminus X|,
\label{eq:pseudo_dist:Passat}
\end{equation*}
 where $\alpha \in [0,1]$. $d^{0}(X,Y) = |Y \setminus X|$ and $d^{1}(X,Y) = |X \setminus Y|$ are respectively the amount of false-negatives and false-positives in $X$ with respect to $Y$. A solution to this minimisation problem can be found by dynamic programming. The details are given in \citep{Passat2011}. In the sequel, we have used the publicly available code of \citet{Naegel2014}.


%
%
\subsection{Experimental validation}
\label{ssec:mat_meth:exp}

We have just established that the LIP-additive homogeneity criterion $H^{\LP}_f(R)$ is theoretically insensitive to a LIP-addition $\LP$ or to a LIP-subtraction $\LM$ of a constant to or from an image, respectively. These operations model the lighting changes caused by a variation of the source intensity or of the exposure-time of the camera. As for the LIP-multiplicative homogeneity criterion $H^{\LT}_f(R)$, it is insensitive to a LIP-multiplication of an image $f$ by a scalar which models the variations of the object thickness or opacity. Let us perform an experimental validation of these insensitivities through simulated and real images. Let us also compare the presented approach to the state-of-the-art method of \citet{Passat2011}.


\subsubsection{Simulated images}
\label{ssec:mat_meth:exp:simu}

For this experiment, we have selected a colour image of a butterfly (Fig.~\ref{fig:simu_var_int:1}) from the dataset YFCC100M (Yahoo Flickr Creative Commons 100M) \citep{Thomee2016,YFCC100M_butterfly2010}. We have extracted its luminance as a grey-level image $f$ (Fig.~\ref{fig:simu_var_int:2}). 
In order to verify the respective insensitivities of the criteria $H^{\LP}$ and $H^{\LT}$ to lighting variations simulated by a LIP-addition $\LP$ or a LIP-multiplication $\LT$, we have darkened and brightened this luminance image $f$ by using each of the LIP-operations.

For the LIP-additive homogeneity criterion $H^{\LP}_f(R)$, a dark image $f_{dk,a}$ is obtained by a LIP-addition of a constant $k=120$ to the complement $f^c$ of the image $f$, $f_{dk,a}= (f^c \LP k)^c$ (Fig.~\ref{fig:simu_var_int:3}). The image complement, which is defined by $f^c = M-1 - f$, allows to take advantage of the LIP-scale. A bright image $f_{br,a}$ is also simulated by a LIP-subtraction of the same constant: $f_{br,a} = (f^c \LM k)^c$ (Fig.~\ref{fig:simu_var_int:4}).

For the LIP-multiplicative homogeneity criterion $H^{\LT}_f(R)$, a dark image $f_{dk,m}$ is obtained by a LIP-multiplication by a scalar $\la_{dr}=4$ of the image complement $f^c$, $f_{dk,m}= (\la_{dr} \LT f^c)^c$ (Fig.~\ref{fig:simu_var_int:5}). A bright image $f_{br,m}$ is also simulated by a LIP-multiplication by a scalar $\la_{br}=0.1$, $f_{br,m} = (\la_{br} \LT f^c)^c$ (Fig.~\ref{fig:simu_var_int:6}).

In all these five images $f$, $f_{dk,a}$ $f_{br,a}$, $f_{dk,m}$ and $f_{br,m}$ the same region $R$ is selected. The homogeneity criteria $H^{\LP}$ and $H^{\LT}$ are then computed in the region $R$ of the images $f$, $f_{dk,a}$, $f_{br,a}$ and of the images $f$, $f_{dk,m}$, $f_{br,m}$, respectively. Their values are given in section ``Results'' (p. \pageref{sec:res}).


\subsubsection{Real images}
\label{ssec:mat_meth:exp:real}

The robustness to lighting variations of the LIP-additive homogeneity criterion $H^{\LP}$ and of the LIP-multiplicative criterion $H^{\LT}$ needs to be verified on real images by using specific experimental material. 

The LIP-additive homogeneity criterion $H^{\LP}$ is expected to have a low sensitivity to variations of source intensity or exposure-time of the camera. In order to verify this assumption, we have conducted an experiment. An image of the same scene is acquired by a camera with significantly different exposure-times and slightly different positions. The scene is composed of a soft toy monster named ``Nessie'' (Fig.~\ref{fig:Nessie}). Three colour images are captured with different exposure-times 1/40~s (Fig.~\ref{fig:Nessie:1}, \ref{fig:Nessie:2}), 1/400~s (Fig.~\ref{fig:Nessie:3}, \ref{fig:Nessie:4}) and 1/800~s (Fig.~\ref{fig:Nessie:3}, \ref{fig:Nessie:5}). They are converted to luminance images in grey levels: $f_0$, $f_1$ and $f_2$, respectively. The shorter the exposure-time is, the darker the image becomes. A segmentation is performed by \citeapos{Revol1997} algorithm using the LIP-additive homogeneity criterion $H^{\LP}$ and a threshold of 200 to decide if a region $R$ is homogeneous (i.e. $H^{\LP}_f(R) \leq 200$) or not. The initialisation of the algorithm is done by a seed point manually selected inside the letters on the body of ``Nessie''. To take advantage of the LIP-scale, the segmentation is performed on the complement of the three luminance images $f_0^c$, $f_1^c$ and $f_2^c$. Three segmented regions $R_0$ (Fig.~\ref{fig:Nessie:1}), $R_1$ (Fig.~\ref{fig:Nessie:3}) and $R_2$ (Fig.~\ref{fig:Nessie:5}) are thereby obtained in the images $f_0$, $f_1$ and $f_2$, respectively. The LIP-additive homogeneity of those regions is then computed. 
For comparison purpose, \citeapos{Passat2011} segmentation method is applied to the images $f_0$ (Fig.~\ref{fig:Nessie:2}), $f_1$ (Fig.~\ref{fig:Nessie:4}) and $f_2$ (Fig.~\ref{fig:Nessie:6}). The user selected regions $G$ are the same seed points which are used for our approach after a dilation by a  square of length side 3
pixels. As the user selected regions $G$ are much smaller than the letters that we want to segment, we choose a parameter $\alpha=0$ in order to minimise the number of false-negatives $d^{0}( \cup_{N \in \mathcal{K}'} N , G )$ in those regions $G$. After minimisation, this number is equal to zero.

In order, to verify the low sensitivity to object opacity of the LIP-multiplicative homogeneity criterion $H^{\LT}$, we have selected three X-ray images of a luggage acquired under different exposures for security purpose (Fig.~\ref{fig:luggage}). The images are coming from a publicly available database, namely GDXray \citep{Mery2015,GDXray}. The object to be detected is a razor blade. Depending on its orientation, the x-rays pass through different thicknesses of the matter and are therefore more or less absorbed.  In the first image $f_0$ (Fig.~\ref{fig:luggage:1}), the razor blade is perpendicular to the x-rays coming from the source. The image of the razor blade is obviously brighter in this image $f_0$ than in the two others $f_1$ (Fig.~\ref{fig:luggage:3}) and $f_2$ (Fig.~\ref{fig:luggage:5}) where the razor blade is oblique to the x-rays. To detect this object, \citeapos{Revol1997} segmentation approach is used with a threshold of 2.7 on the LIP-multiplicative homogeneity criterion $H^{\LT}$ of a region. The algorithm is initialised by manually selected seed points inside the razor blade (Fig.~\ref{fig:luggage:1}, \ref{fig:luggage:3} and \ref{fig:luggage:5}). Three regions $R_0$ (Fig.~\ref{fig:luggage:2}), $R_1$ (Fig.~\ref{fig:luggage:3}) and $R_2$ (Fig.~\ref{fig:luggage:5}) corresponding to the object are then obtained in the images $f_0$, $f_1$ and $f_2$, respectively. The LIP-multiplicative homogeneity of those regions is then estimated. Its values are presented in the following section.

%
%
\section{Results}
\label{sec:res}

The experimental results are given for the simulated and the real images.


\subsection{Simulated images}
\label{ssec:res:simu}

The LIP-additive homogeneity criterion $H^{\LP}$ of the region is computed for the image $f$ (Fig.~\ref{fig:simu_var_int:2}), its darkened version $f_{dk,a}$ by LIP-addition (Fig.~\ref{fig:simu_var_int:3}) and its brightened version $f_{br,a}$ by LIP-subtraction (Fig.~\ref{fig:simu_var_int:4}). For each image, the criterion $H^{\LP}$ of the region $R$ is equal to:\\
$H^{\LP}_{f^c}(R) = H^{\LP}_{f^{c}_{dk,a}}(R) = H^{\LP}_{f^{c}_{br,a}}(R) = 150.7$. 
This result verifies thereby the insensitivity of the LIP-additive homogeneity criterion $H^{\LP}$ to LIP-addition $\LP$ or LIP-subtraction $\LM$ of a constant to or from an image, respectively.

The LIP-multiplicative homogeneity criterion $H^{\LT}$ is also estimated in the image $f$ (Fig.~\ref{fig:simu_var_int:2}) and its darkened and brightened versions by LIP-multiplication $f_{dk,m}$ (Fig.~\ref{fig:simu_var_int:5}) and $f_{br,m}$ (Fig.~\ref{fig:simu_var_int:6}), respectively. For each image the criterion $H^{\LT}$ of the region $R$ is equal to:\\
$H^{\LT}_{f^c}(R) = H^{\LT}_{f^{c}_{dk,m}}(R) = H^{\LT}_{f^{c}_{br,m}}(R) = 3.33$. This result shows the insensitivity of the LIP-multiplicative criterion $H^{\LT}$ to the LIP-multiplication $\LT$ of an image by a scalar.

\begin{figure}[!t]
\centering
\subfloat[]{\includegraphics[width=4cm]{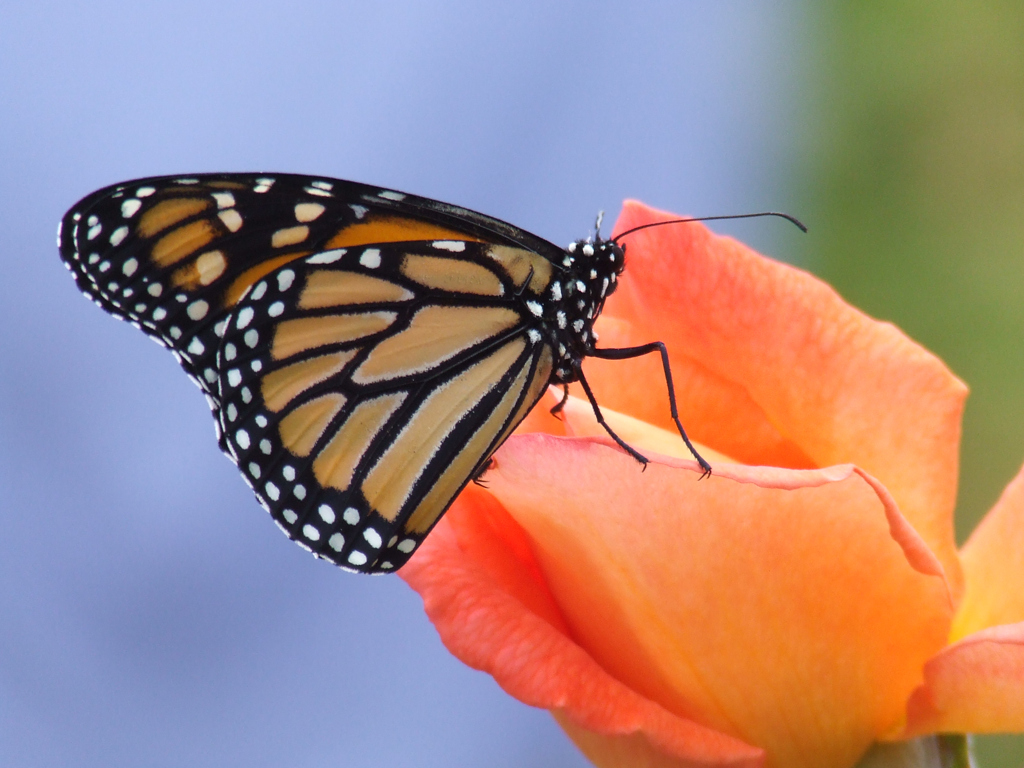}%
\label{fig:simu_var_int:1}}
\hfill
\subfloat[$f$]{\includegraphics[width=4cm]{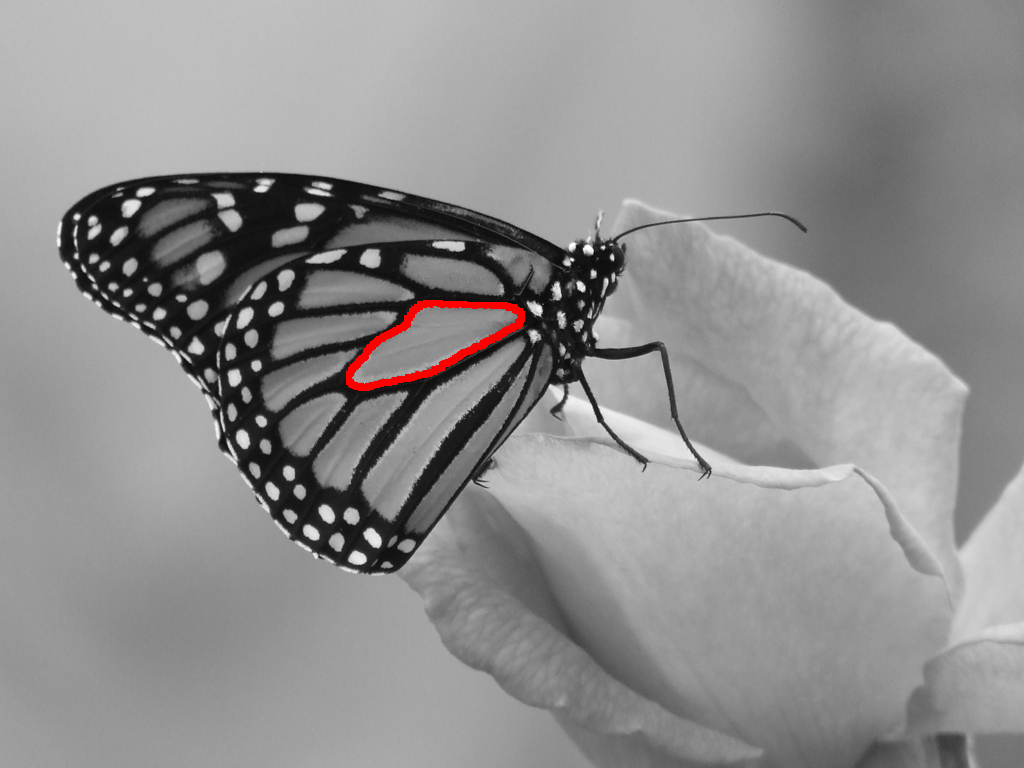}%
\label{fig:simu_var_int:2}}
\hfill
\subfloat[$f_{dk,a} = (f^c \protect \LP 120)^c$]{\includegraphics[width=4cm]{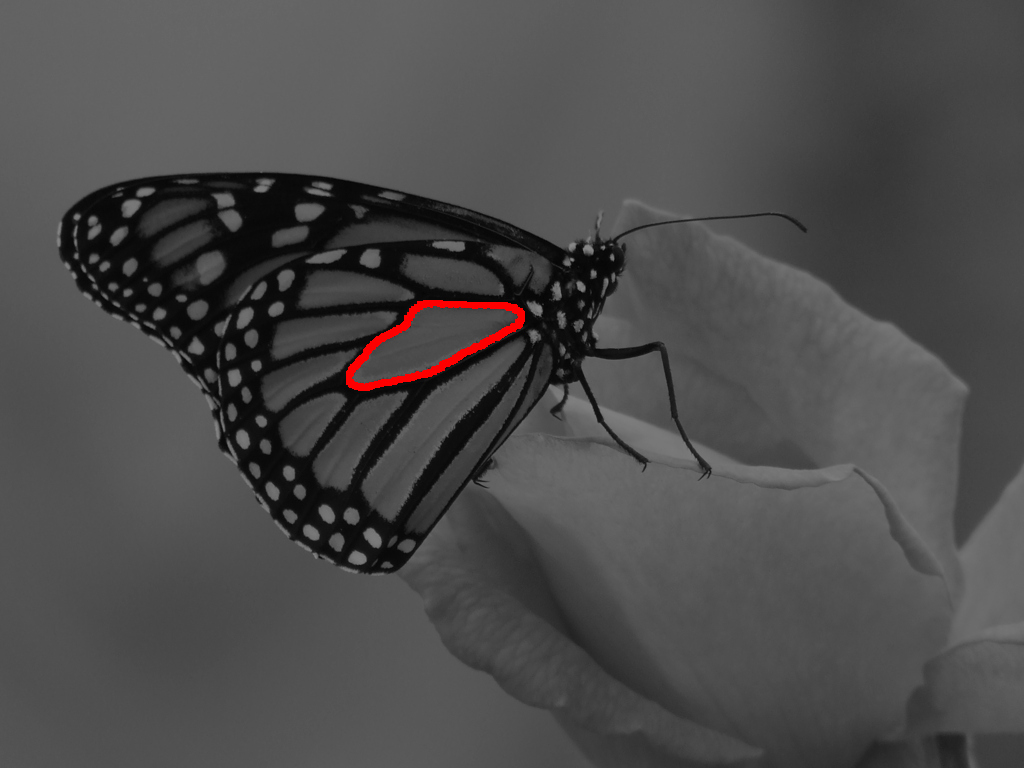}%
\label{fig:simu_var_int:3}}
\hfill
\subfloat[$f_{br,a} = (f^c \protect \LM 120)^c$]{\includegraphics[width=4cm]{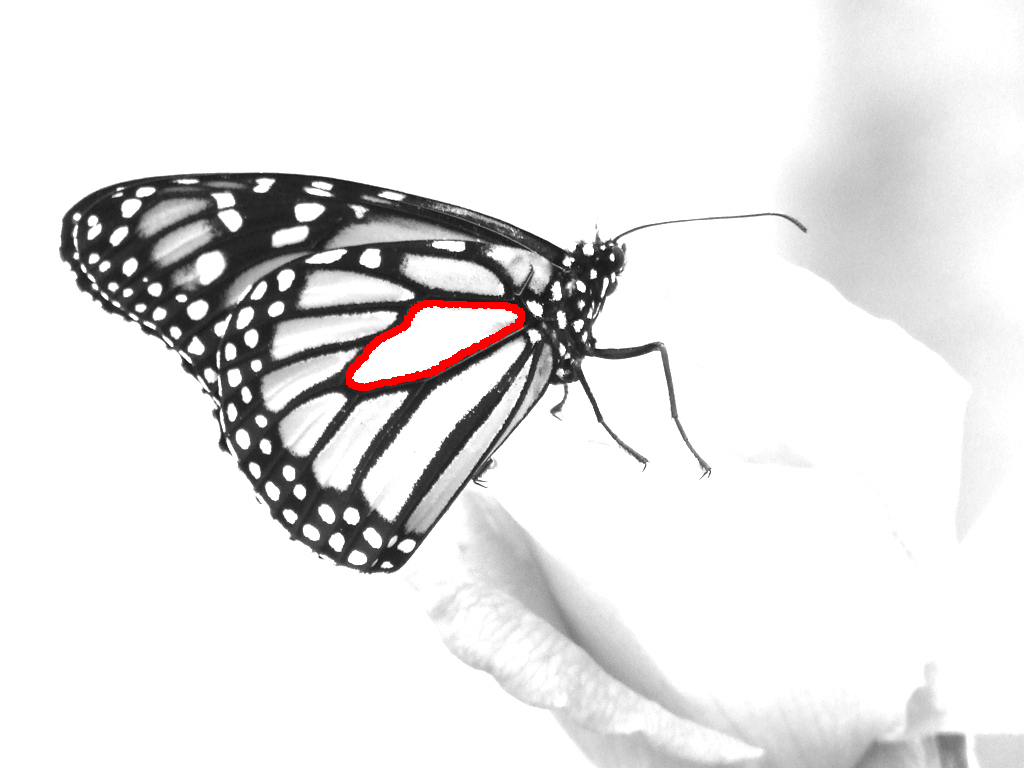}%
\label{fig:simu_var_int:4}}
\hfill
\subfloat[$f_{dk,m}= (4 \protect \LT f^c)^c$]{\includegraphics[width=4cm]{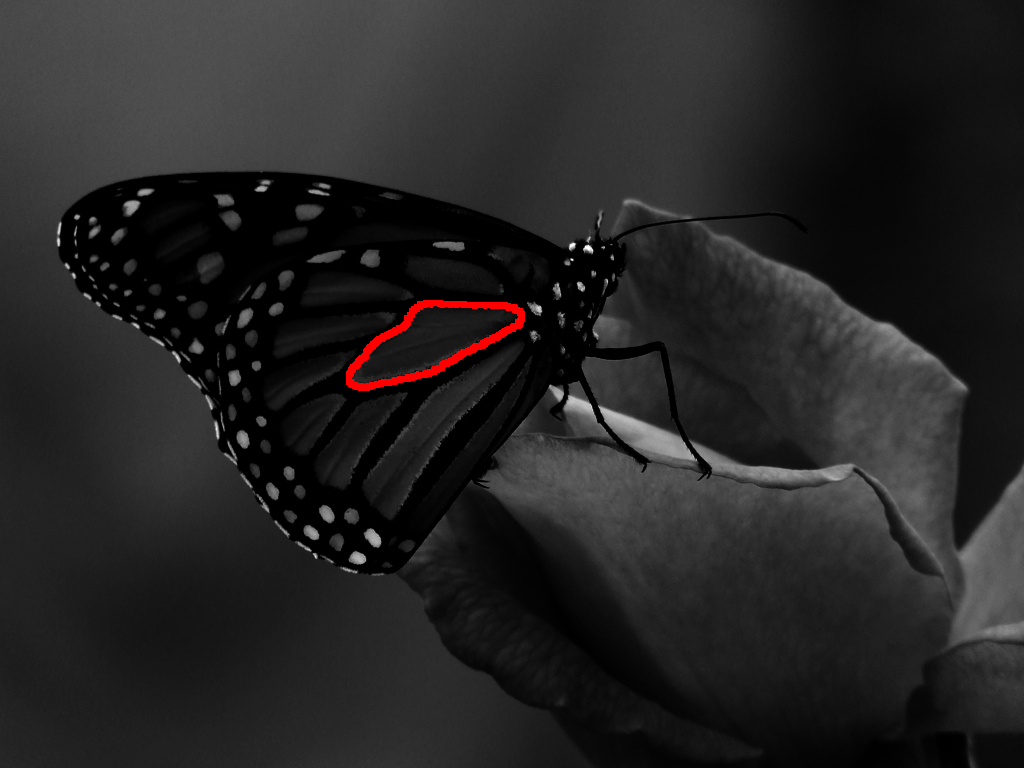}%
\label{fig:simu_var_int:5}}
\hfill
\subfloat[$f_{br,m} = (0.1 \protect \LT f^c)^c$]{\includegraphics[width=4cm]{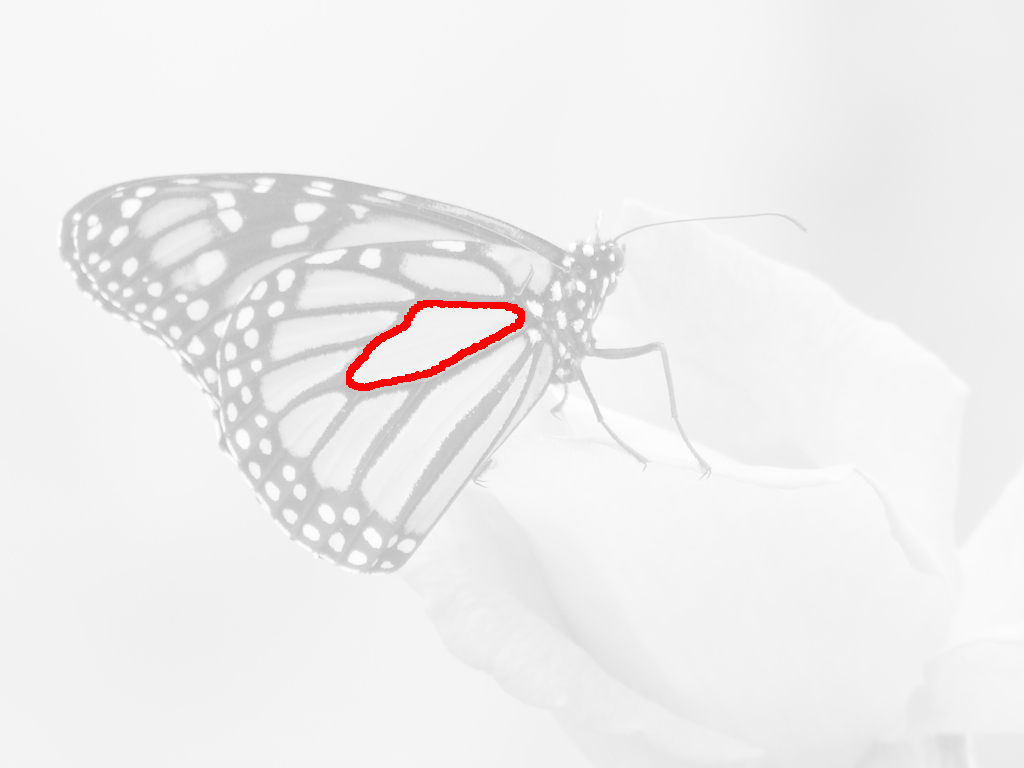}%
\label{fig:simu_var_int:6}}
\caption{(a) Colour image of a butterfly and (b) its luminance image $f$ with a region $R$ in red. (c) Darkened image $f_{dk,a}$ obtained by LIP-addition of a constant $k=120$. (d) Brightened image $f_{br,a}$ obtained by LIP-subtraction of $k$. 
(e) Darkened image $f_{dk,m}$ obtained by a LIP-multiplication of $f^c$ by $\la_{dr}=4$. (f) Brightened image $f_{br,m}$ obtained by a LIP-multiplication by $\la_{br}=0.1$.}
\label{fig:simu_var_int}
\end{figure}

\subsection{Real images}
\label{ssec:res:real}

Let us give the results obtained on the real images of figures \ref{fig:Nessie} and \ref{fig:luggage}.

The LIP-additive homogeneity criterion $H^{\LP}$ is illustrated in figure \ref{fig:Nessie}. One can notice that the segmentations of the letters on the body of ``Nessie'' are very similar in all the three images $f_0$ (Fig.~\ref{fig:Nessie:1}), $f_1$ (Fig.~\ref{fig:Nessie:3}) and $f_2$ (Fig.~\ref{fig:Nessie:5}) using this criterion $H^{\LP}$ within the \citeapos{Revol1997} method (i.e. our approach). In addition, the values of the homogeneity criterion are as follows: $H^{\protect \LP}_{f_0^c}(R_0) = 198.5$, $H^{\protect \LP}_{f_1^c}(R_1) = 197.5$ and $H^{\protect \LP}_{f_2^c}(R_2) = 197.0$. These values are very close for the three regions $R_0$, $R_1$ and $R_2$ of these images which were acquired with significantly different exposure-times. Those results prove therefore the robustness of the LIP-additive homogeneity criterion $H^{\LP}$ in \citeapos{Revol1997} segmentation method to the variations of the camera exposure-time. 
This approach is also compared to the one of \citet{Passat2011}. One can notice that the segmentations with this latter method are different between the three images $f_0$ (Fig.~\ref{fig:Nessie:2}), $f_1$ (Fig.~\ref{fig:Nessie:4}) and $f_2$ (Fig.~\ref{fig:Nessie:6}). Those results show that \citeapos{Passat2011} method lacks of robustness for important variations of the camera exposure-time, whereas ours is robust to such variations. In addition, the letters are not entirely extracted with \citeapos{Passat2011} method whereas with ours, their segmentations are better (Fig. \ref{fig:Nessie:1}, \ref{fig:Nessie:3} and \ref{fig:Nessie:5}).

Figure \ref{fig:luggage} presents the results obtained for the LIP-multiplicative criterion $H^{\LT}$. The segmentations of the razor blade are very similar in the three images (Fig.~\ref{fig:luggage:2}, \ref{fig:luggage:4} and \ref{fig:luggage:6}) using this criterion $H^{\LT}$ within the \citeapos{Revol1997} method. As the three images were captured with different exposures, the x-rays were differently absorbed by the object. In addition, the values of the homogeneity criterion are the following for each segmented regions: $H^{\protect \LT}_{f_0^c}(R_0) = 2.11$, $H^{\protect \LT}_{f_1^c}(R_1) = 2.50$ and $H^{\protect \LT}_{f_2^c}(R_2) = 1.94$. One can notice that the criterion values are of similar amplitudes for the three regions of an object presenting different thicknesses which absorb the x-rays. 
This experiment illustrates the robustness of the LIP-multiplicative criterion $H^{\LT}$ in \citeapos{Revol1997} segmentation method to different object opacities (or absorptions).

%
%
\section{Discussion}

The results obtained with the simulated images show that the LIP-additive homogeneity criterion $H^{\LP}$ is insensitive to the LIP-addition $\LP$ or the LIP-subtraction $\LM$ of the image complement $f^c$ by a constant. These operations $\LP$ and $\LM$ model a darkening or a brightening of the image by a variation of the source intensity or the exposure-time of the camera \citep{Jourlin2016}. This result proves the theoretical insensitivity of the LIP-additive criterion $H^{\LP}$ to such variations. Similarly, the LIP-multiplicative homogeneity criterion $H^{\LT}$ is theoretically insensitive to the variations of the object opacity or thickness which are modelled by a LIP-multiplication $\LT$ by a scalar \citep{Jourlin2016}.

\begin{figure}[!t]
\centering
\subfloat[$f_0$, $H^{\protect \LP}_{f_0^c}(R_0) = 198.5$]{\includegraphics[width=4cm]{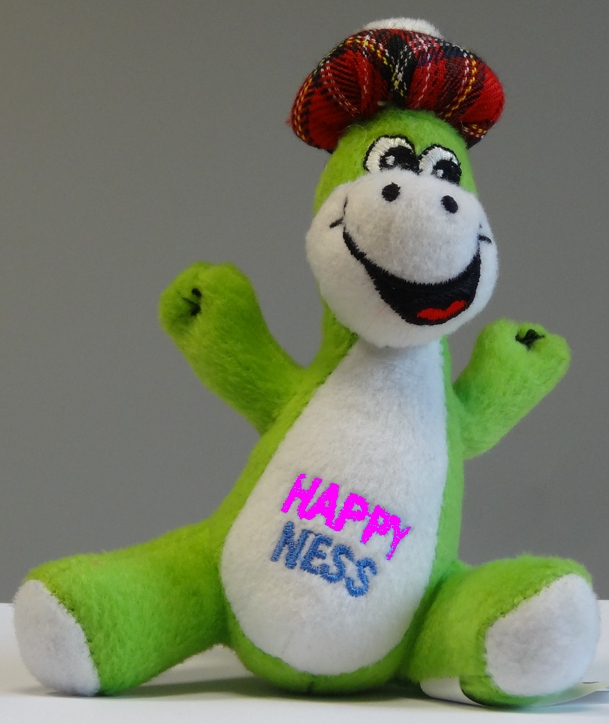}%
\label{fig:Nessie:1}}
\hfill
\subfloat[$f_0$, \citet{Passat2011}]{\includegraphics[width=4cm]{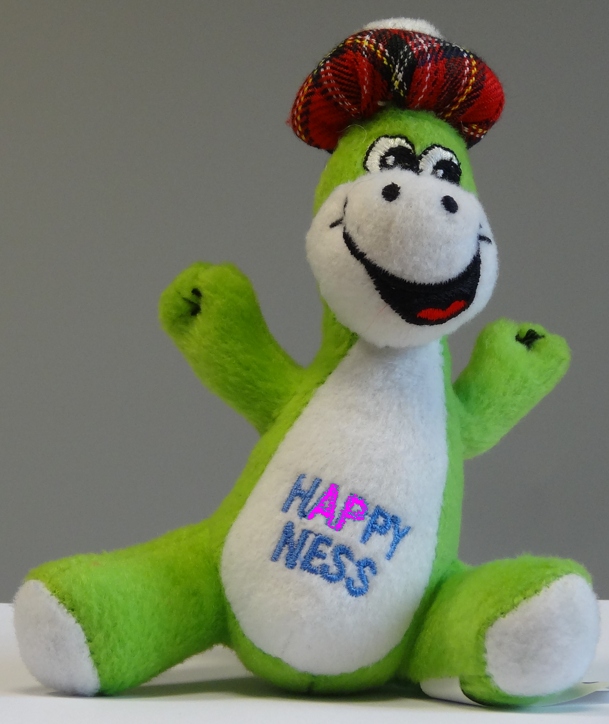}%
\label{fig:Nessie:2}}\\

\subfloat[$f_1$, $H^{\protect \LP}_{f_1^c}(R_1) = 197.5$]{\includegraphics[width=4cm]{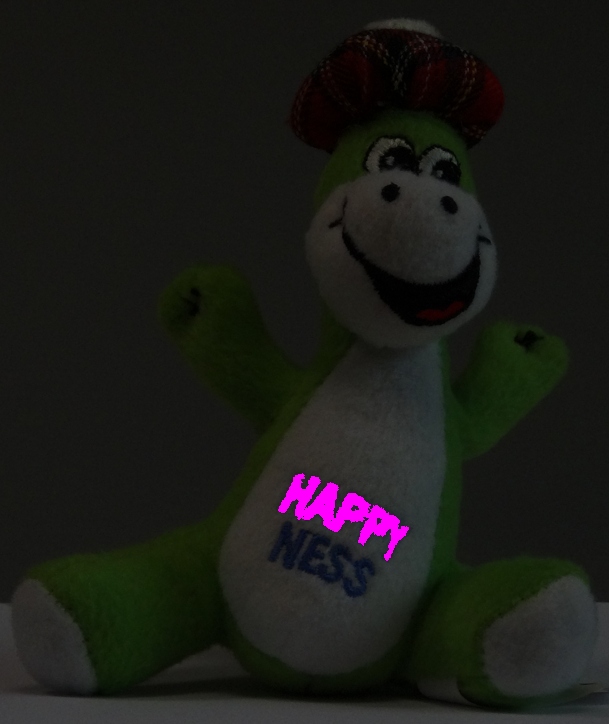}%
\label{fig:Nessie:3}}
\hfill
\subfloat[$f_1$, \citet{Passat2011}]{\includegraphics[width=4cm]{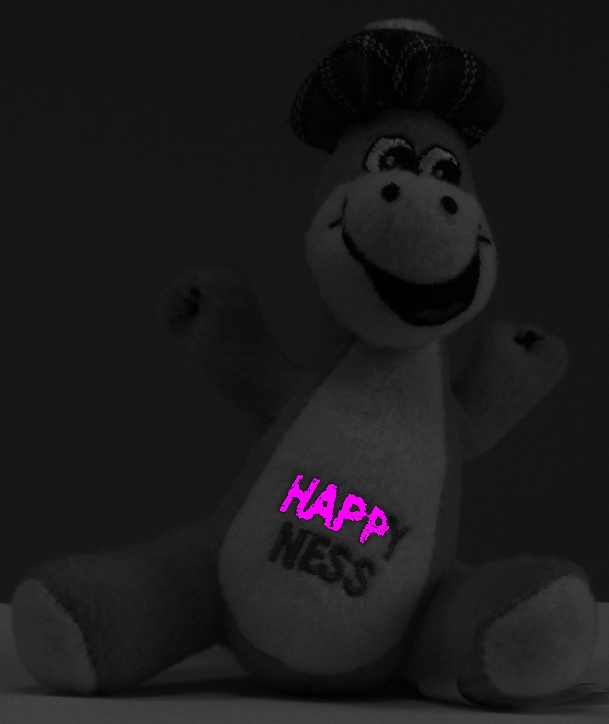}%
\label{fig:Nessie:4}}\\

\subfloat[$f_2$, $H^{\protect \LP}_{f_2^c}(R_2) = 197.0$]{\includegraphics[width=4cm]{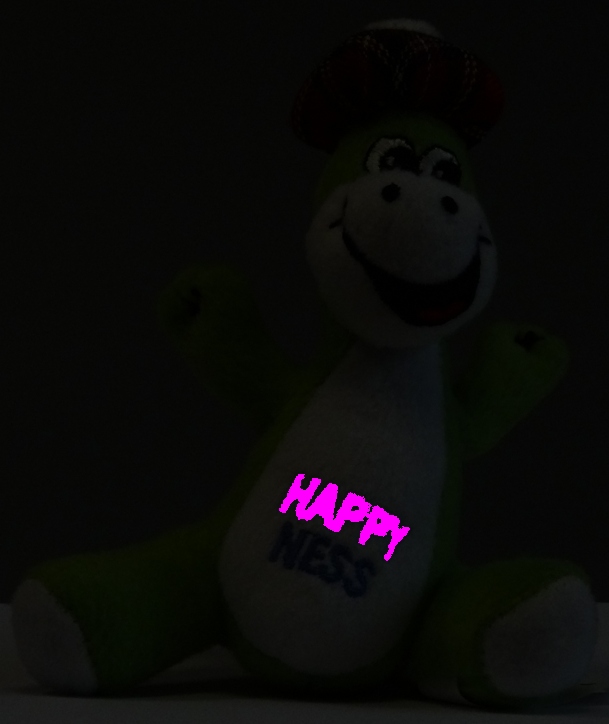}%
\label{fig:Nessie:5}}
\hfill
\subfloat[$f_2$, \citet{Passat2011}]{\includegraphics[width=4cm]{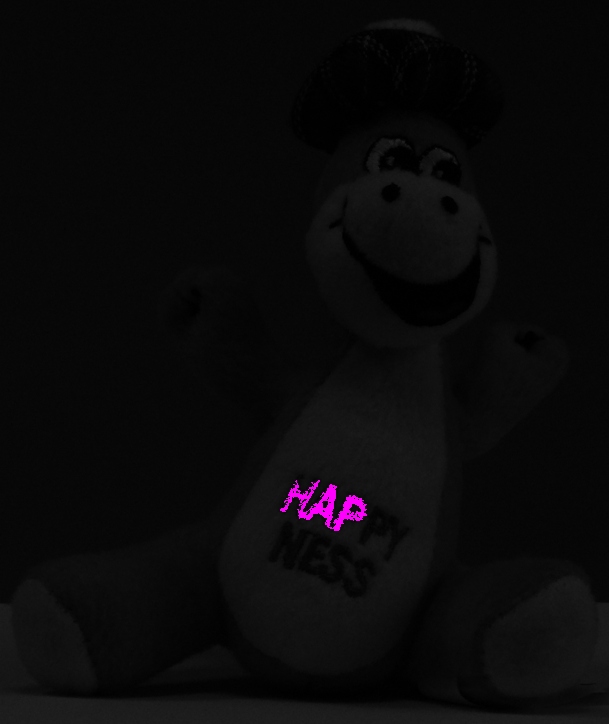}%
\label{fig:Nessie:6}}
\caption{Comparison of the robustness to three exposure-time variations (a, c, e) of \citeapos{Revol1997} method with a LIP-additive homogeneity criterion $H^{\protect \LP}$ and (b, d, f) of \citeapos{Passat2011} method. Colour versions of the luminance images (a, b) $f_0$, (c, d) $f_1$ and (e, f) $f_2$ acquired at 1/40~s, 1/400~s and 1/800~s, respectively. The segmented regions are depicted in magenta colour. The homogeneity criterion values $H^{\protect \LP}_{f_0^c}(R_0)$, $H^{\protect \LP}_{f_1^c}(R_1)$, $H^{\protect \LP}_{f_2^c}(R_2)$ of the regions $R_0$, $R_1$ and $R_2$ are given under the images (a), (c) and (e), respectively.}
\label{fig:Nessie}
\end{figure}

\pagebreak

\begin{figure}[t]
\centering
\subfloat[$f_0$]{\includegraphics[width=4cm]{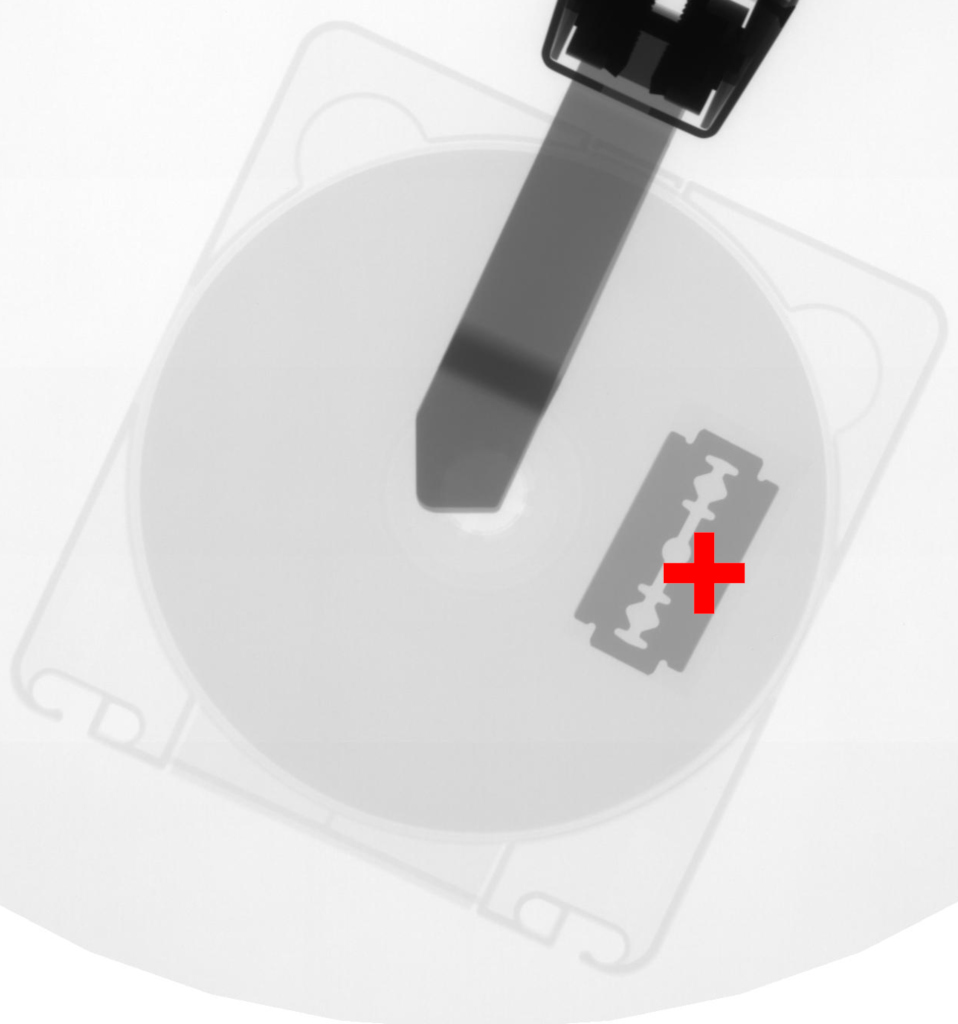}%
\label{fig:luggage:1}}
\hfill
\subfloat[$H^{\protect \LT}_{f_0^c}(R_0) = 2.11$]{\includegraphics[width=4cm]{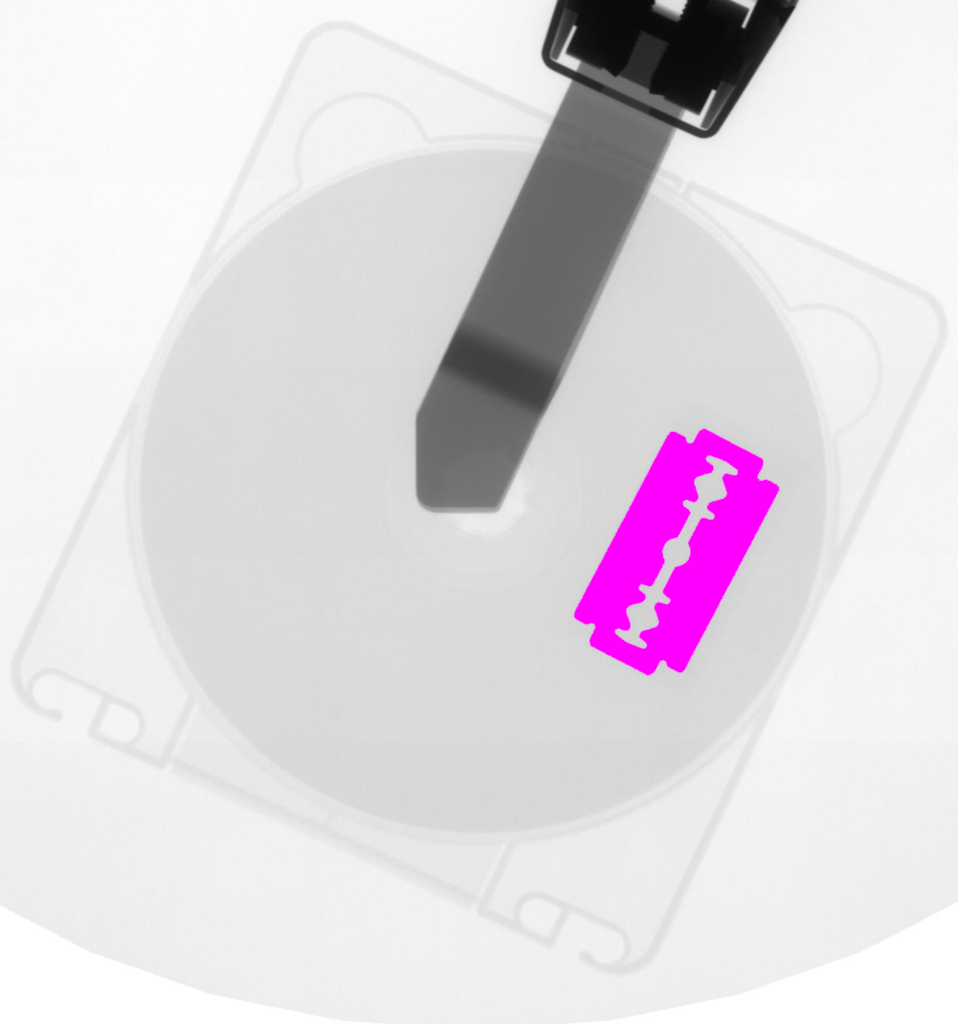}%
\label{fig:luggage:2}}\\
\subfloat[$f_1$]{\includegraphics[width=4cm]{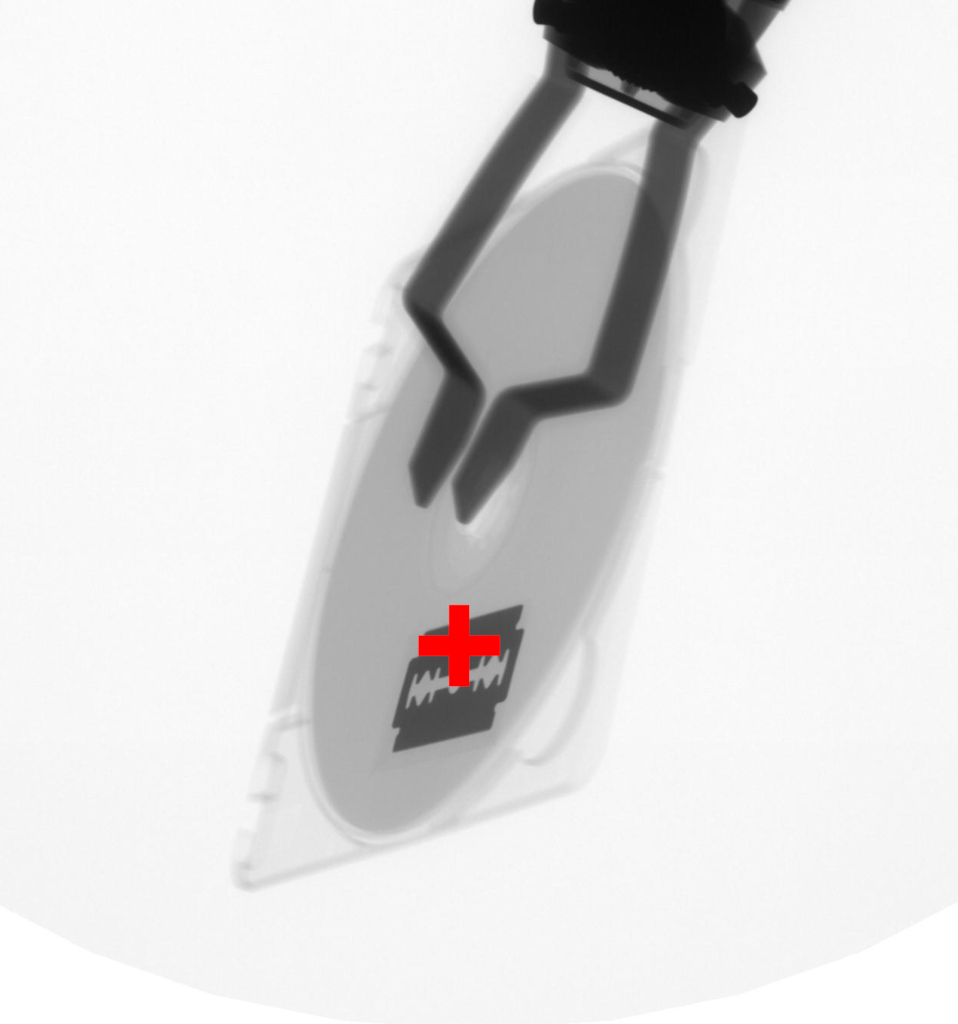}%
\label{fig:luggage:3}}
\hfill
\subfloat[$H^{\protect \LT}_{f_1^c}(R_1) = 2.50$]{\includegraphics[width=4cm]{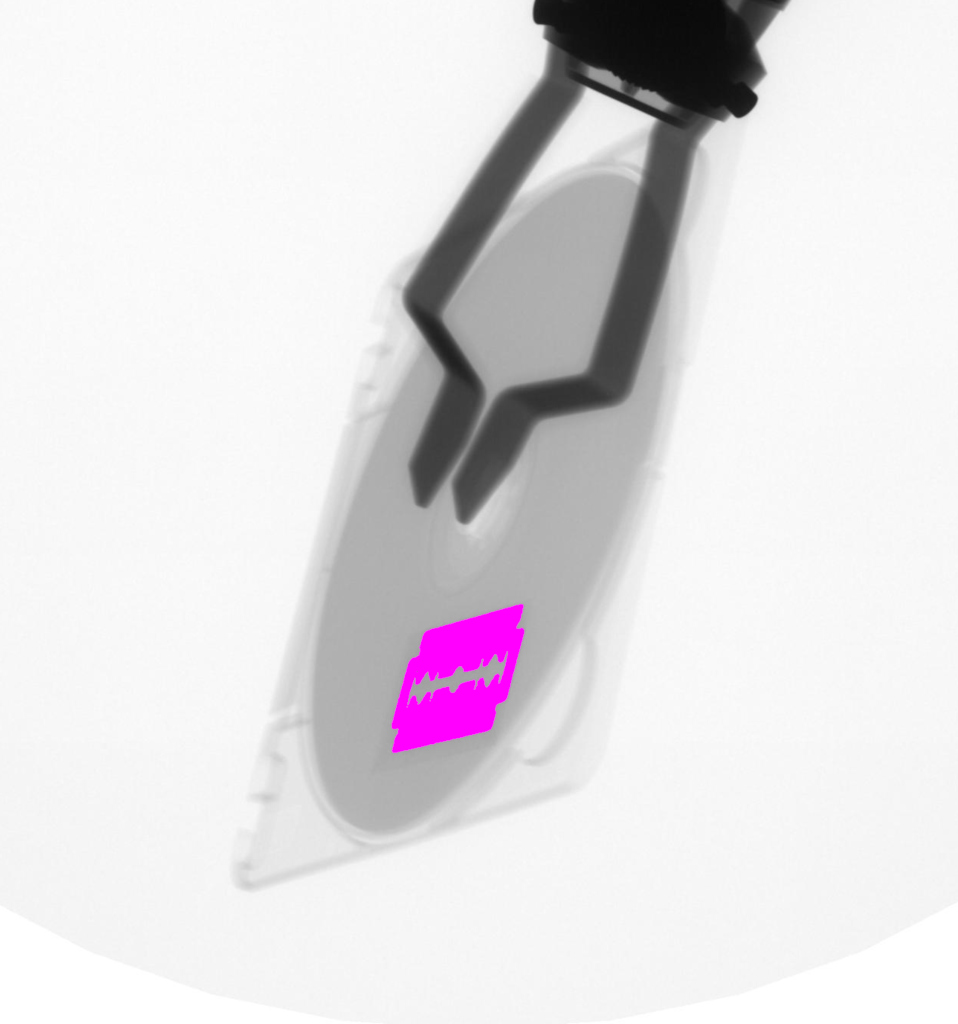}%
\label{fig:luggage:4}}\\
\subfloat[$f_2$]{\includegraphics[width=4cm]{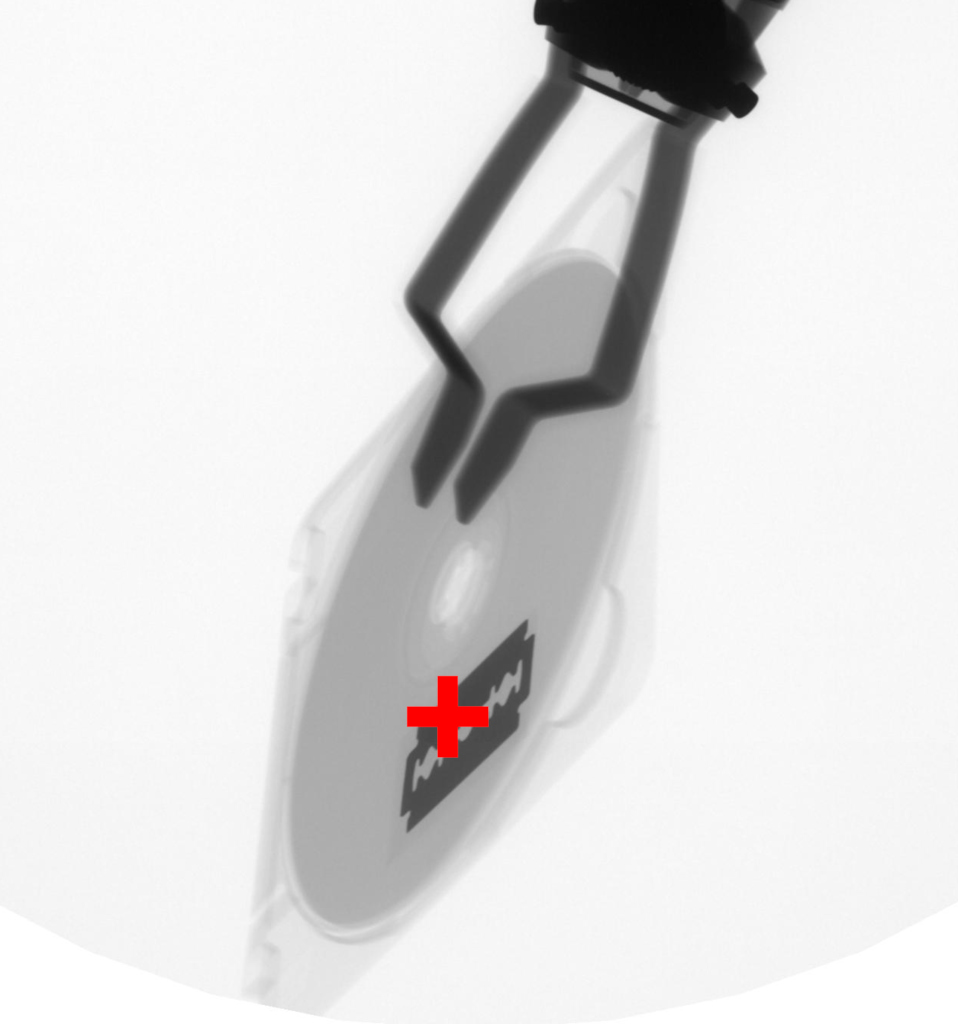}%
\label{fig:luggage:5}}
\hfill
\subfloat[$H^{\protect \LT}_{f_2^c}(R_2) = 1.94$]{\includegraphics[width=4cm]{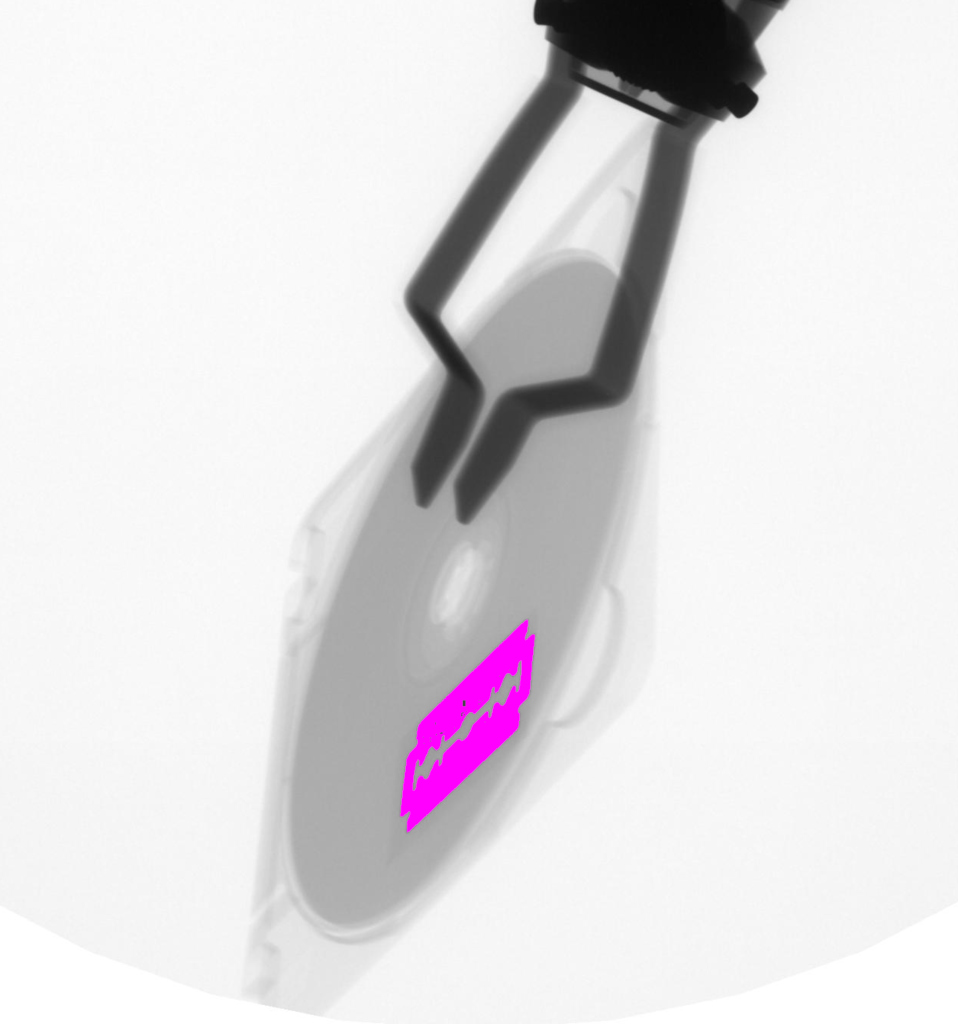}%
\label{fig:luggage:6}}
\caption{Robustness to variations of object opacity of the segmentation by LIP-multiplicative homogeneity criterion $H^{\protect \LT}$. (a) (b) and (c) Three images $f_0$, $f_1$ and $f_2$ acquired for different exposures of a luggage containing a razor blade. The seed points are shown by a red cross. (d) (e) and (f). The segmented regions $R_0$, $R_1$ and $R_2$ of the images $f_0$, $f_1$ and $f_2$, respectively, are shown in magenta colour. The values of the homogeneity criteria $H^{\protect \LT}_{f_0^c}(R_0)$, $H^{\protect \LT}_{f_1^c}(R_1)$, $H^{\protect \LT}_{f_2^c}(R_2)$ are given under the images (b), (d) and (f).}
\label{fig:luggage}
\end{figure}

These simulations are confirmed by the experimental results. Figure \ref{fig:Nessie} proves the robustness of the LIP-additive homogeneity criterion $H^{\LP}_f(R)$ to lighting changes caused by a variation of an exposure-time of the camera. This lighting variation is equivalent to a variation of the source intensity.
Although the methods based on levels sets are theoretically robust to a contrast change caused by a continuous and increasing function applied to the image, the results obtained in figure \ref{fig:Nessie} prove that the level set approach of \citet{Passat2011} lacks of robustness for important variations of the camera exposure-time.
The results given in figure \ref{fig:luggage} prove the robustness of the LIP-multiplicative homogeneity criterion $H^{\LT}_f(R)$ to lighting changes caused by variations of the object opacity.


The previous results show that the level sets methods based on component-trees \citep{Salembier1998,Passat2011} or on inclusion trees \citep{Monasse2000,Xu2016} will therefore strengthen their robustness to strong contrast changes - due to an optical cause - by using a LIP homogeneity criterion. The optical cause can either be a variation of the source intensity, which will be corrected by the LIP-additive criterion $H^{\LP}$, or a variation of the object opacity, which will be corrected by the LIP-multiplicative criterion $H^{\LT}$. These findings will be studied in a future paper.
%
%
\section{Conclusion}

We have therefore successfully introduced two new region homogeneity criteria which are robust to lighting changes, namely the LIP-additive homogeneity criterion $H^{\LP}$ and the LIP-multiplicative criterion $H^{\LT}$. With experiments on simulated and on real images, we have shown that the LIP-additive homogeneity criterion $H^{\LP}$ is robust to changes caused by variations of source intensity (or exposure-time of the camera) whereas the LIP-multiplicative criterion $H^{\LT}$ is robust to changes due to variations of object opacity (or thickness). The introduction of those criteria in \citeapos{Revol1997} segmentation method gives it the same robustness. 
Compared to \citeapos{Passat2011} method based on the component-tree of the image level sets, ours is more robust to strong intensity changes due to one of the previous optical cause. As far as we know, this is the first time that a segmentation method based on a region homogeneity parameter robust to those type of lighting changes has been shown. This novel method of segmentation paves the way to numerous applications presenting strong lighting variations.

%
%

\section{Acknowledgements}

The authors thanks the International Prevention Research Institute for its constant support for this work. They are also grateful to Dr. Eileen Boyle for her careful re-reading of the manuscript.

%
%

\bibliography{refs}

\end{paper}
\end{document}